\documentclass[10pt,twocolumn,letterpaper]{article}

\usepackage{authblk}
\usepackage{cvpr}              
\usepackage[accsupp]{axessibility} 
\usepackage{graphicx}
\usepackage{amsmath}
\usepackage{amssymb}
\usepackage{booktabs}

\usepackage{graphicx}
\usepackage{multirow}
\newtheorem{thm}{Theorem}[section]
\newtheorem{mydef}[thm]{Definition}
\newtheorem{mycor}[thm]{Corollary}

\newtheorem{mylem}[thm]{Lemma}
\newtheorem{myrem}{Remark}

\newtheorem{innercustomthm}{Proof}
\newenvironment{custompro}[1]
  {\renewcommand\theinnercustomthm{#1}\innercustomthm}
  {\endinnercustomthm}

\usepackage{caption}

\usepackage{varwidth}
\DeclareCaptionFormat{myformat}{%
  \begin{varwidth}{\linewidth}%
    \centering
    #1#2#3%
  \end{varwidth}%
}

\usepackage[pagebackref,breaklinks,colorlinks]{hyperref}

\usepackage[capitalize]{cleveref}
\crefname{section}{Sec.}{Secs.}
\Crefname{section}{Section}{Sections}
\Crefname{table}{Table}{Tables}
\crefname{table}{Tab.}{Tabs.}

\renewcommand{\vec}[1]{\mathbf{#1}}

\newcommand{\x}{\vec{x}}
\newcommand{\y}{\vec{y}}

\newcommand{\D}{\mathcal{D}}
\newcommand{\U}{\vec{U}}
\newcommand{\w}{\vec{w}}

\newcommand{\HH}{\vec{H}}

\newcommand{\A}{\vec{A}}
\newcommand{\s}{\vec{s}}
\newcommand{\St}{\mathcal{S}}

\newcommand{\h}{\vec{h}}
\newcommand{\hh}{h^2}
\newcommand{\ac}{\vec{a}}
\newcommand{\W}{\vec{W}}

\newcommand{\KL}{\mathrm{KL}}
\newcommand{\g}{\vec{g}}

\newcommand{\act}{\mathrm{\vec{act}}}
\newcommand{\Z}{\mathcal{Z}}

\newcommand{\T}{\intercal}

\newcommand{\Pro}{\mathbb{P}}
\newcommand{\E}{\mathbb{E}}
\newcommand{\N}{\mathbb{N}}
\newcommand{\R}{\vec{R}}
\newcommand{\Lc}{\mathcal{L}}

\newcommand{\V}{\mathcal{V}}

\newcommand{\vecc}{\mathrm{vec}}
\newcommand{\lip}{\mathrm{lip}}
\newcommand{\fgm}{\mathrm{fgm}}

\newcommand{\so}{S$^2$O}
\newcommand{\adv}{\rm{adv}}

\newcommand{\ul}{\vec{u}}

\newcommand{\commentout}[1]{}

\newcommand{\gaojie}[1]{{\color{cyan}{}#1}}

\def\cvprPaperID{4422} 
\def\confName{CVPR}
\def\confYear{2022}

\begin{document}

\title{Enhancing Adversarial Training with Second-Order Statistics of Weights}

\author[1]{Gaojie Jin}
\author[2]{Xinping Yi}
\author[1]{Wei Huang}
\author[1]{Sven Schewe}
\author[1]{Xiaowei Huang \thanks{Corresponding author}}
\affil{ Department of Computer Science, University of Liverpool, Liverpool, UK}
\affil[2]{Department of Electrical Engineering and Electronics, University of Liverpool, Liverpool, UK \authorcr
\{g.jin3, xinping.yi, w.huang23, svens, xiaowei.huang\}@liverpool.ac.uk}
\renewcommand*{\Authands}{, }

\maketitle

\begin{abstract}
Adversarial training has been shown to be one of the most effective approaches to improve the robustness of deep neural networks.
It is 
formalized as a min-max optimization over model weights and adversarial perturbations, where the weights can be optimized through gradient descent methods like SGD.
In this paper, we show that treating model weights as random variables allows for enhancing adversarial training through  \textbf{S}econd-Order \textbf{S}tatistics \textbf{O}ptimization (S$^2$O) with respect to the weights.
By relaxing a common (but unrealistic) assumption of previous PAC-Bayesian frameworks that all weights are statistically independent, 
we 
derive
an improved PAC-Bayesian adversarial generalization bound, which suggests that optimizing second-order statistics of weights can effectively tighten the bound.
In addition to this theoretical insight, we conduct an extensive set of experiments, which show that S$^2$O not only improves the robustness and  generalization of the trained neural networks when used in isolation, but also integrates easily in state-of-the-art adversarial training techniques like TRADES, AWP, MART, and AVMixup, leading to a measurable improvement of these techniques. The code is available at \url{https://github.com/Alexkael/S2O}.
\end{abstract}

\vspace{-3mm}
\section{Introduction}
\vspace{-1mm}
It is well known that it is simple to fool convolutional neural networks -- to whom we refer as neural networks in this paper -- to make incorrect predictions with high confidence by adding 
human-imperceptible perturbations to their input
\cite{goodfellow2014explaining,szegedy2013intriguing,wu2020skip,mu2021sparse}.
Among many different approaches \cite{bai2019hilbert,ma2018characterizing,tramer2017ensemble,xu2017feature,papernot2016distillation} to detect or reduce such adversarial examples, adversarial training~\cite{papernot2016distillation,madry2017towards} is known to be the most effective~\cite{athalye2018obfuscated}. 


\begin{figure*}[t!]
\includegraphics[width=1
\textwidth]{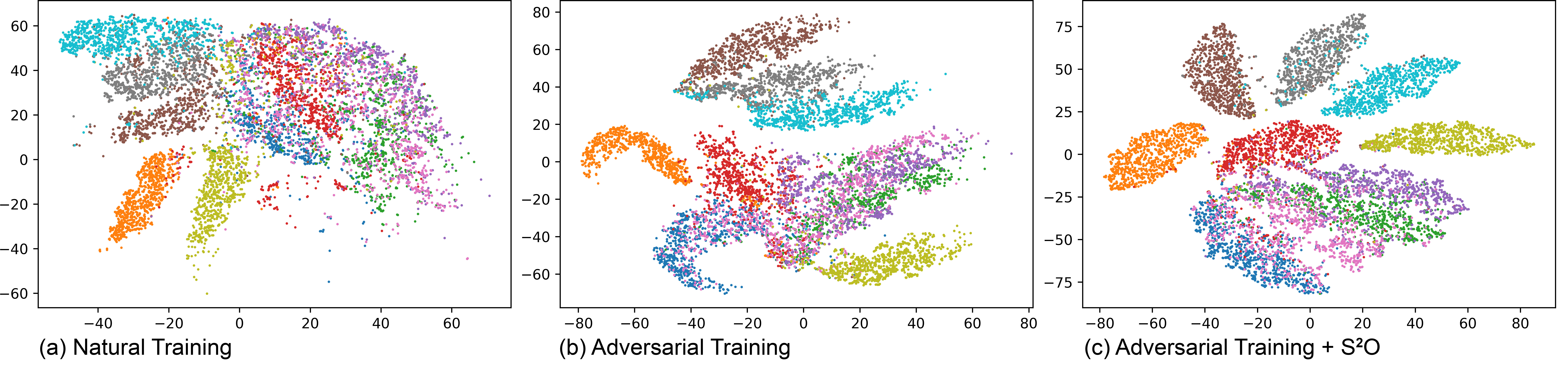}
\centering
\vspace{-5mm}
\caption{Visualization of the (PGD-20) distributions of penultimate latent representations for each label through t-SNE with trained CNN models on Fashion-MNIST. (\textbf{a}) natural training; (\textbf{b}) adversarial training (PGD-10); and (\textbf{c}) adversarial training (PGD-10) with \so.}   
\vspace{-4mm}
\label{fig:1}
\end{figure*}

Adversarial training is 
formulated as a min-max optimization problem, where the inner maximization is to find the worst-case adversarial perturbations for the training instances, while the outer minimization is to reduce the loss induced by these adversarial perturbations. 
%
While finding the optimal solution to this min-max optimization is challenging,
the current wisdom 
is to first decompose the min-max problem into a master minimization problem and a slave maximization problem, and to then solve them via alternating optimization.
Both, the minimization and the maximization, are typically solved by utilizing the gradients of the loss function $\Lc$. 
In particular, the master minimization updates the weights according to the gradient over the weight, i.e., $\nabla_{\textbf{W}}\Lc$.

%

This paper takes a drastically different view:
we believe that adversarial training can benefit from also considering a second-order statistics over weight.  
Our study covers both theoretical and empirical perspectives. 

Our theoretical argument is obtained through updating the PAC-Bayesian framework \cite{mcallester1999pac,dziugaite2017computing}, which only deals with the model generalization in its original format, by considering adversarial robustness and a second order statistics over weight. Under Bayesian regime, weights are random variables, and a model is a sample drawn from an \emph{a posteriori} distribution. 
Our update of the framework draws from two aspects as described in \cref{sec:bound}. 
First, by relaxing the unreasonable assumption that all weights are statistically independent, 
we 
introduce a second-order statistics of weights, i.e., a weight correlation matrix (or normalized covariance matrix).
Second, as in \cite{farnia2018generalizable}, we consider the adversarial robustness in addition to the model generalization.

This updated framework provides a theoretical indication that we may  
control an adversarial generalization bound during training -- and thus improve both robustness and generalization of the resulting model -- by monitoring some norms (\emph{e.g.,} singular value, spectral norm, determinant) of the weight correlation matrix. 
To enable such a control, we need methods to estimate the weight correlation matrix and conduct training, respectively. 
As described 
in \cref{sec:estimate} we employ two methods for the former: 
one is a sampling method  and the other a Laplace approximation method inspired by \cite{botev2017practical,ritter2018scalable}.
For the latter, 
we propose a novel \textbf{S}econd-Order \textbf{S}tatistics \textbf{O}ptimization (\so) method. 

To intuitively understand why \so\ can improve the robustness, \cref{fig:1} shows that adversarial training based on \so\ (abbreviated AT+\so) leads to a visually improved separation between points of different classes under PGD-20 attack -– a clear sign of increased robustness. 

Through extensive experiments and comparison with the strate-of-the-art in \cref{sec:experiments}, we show that S$^2$O can not only significantly improve the robustness and generalization of the trained models by itself, but also enhance the existing adversarial training techniques further.
Notably, \so\ can be used in a plug-and-play manner with other 
adversarial training techniques,
including TRADES\cite{zhang2019theoretically}, MART\cite{wang2019improving}, AWP\cite{wu2020adversarial}, and AVMixup\cite{lee2020adversarial}, four state-of-the-art adversarial training techniques that represent different directions of the effort to improve the min-max optimization scheme.
Importantly, we note that the enhancement from \so\
only leads to a marginal increase of the training time (\cref{sec:5.2}). 

We remark that \so\ is different from other second-order adversarial training methods: \emph{e.g.,} \cite{ma2020soar} proposes an adversarial regularization through approximating the
loss function as a second-order Taylor series expansion, \cite{tsiligkaridis2020second} improves robustness via Hessian matrix of the input, and \cite{singla2020second} studies adversarial attack through Hessian of the weight matrix (while we focus on the weight correlation matrix). It is also different from the ``\emph{weight orthogonality}'' 
in \cite{saxe2013exact,mishkin2015all,bansal2018can}: 
while the weight correlation matrix is a statistical property of weight matrices treated as random variables, orthogonality treats weight matrices as deterministic variables.


\section{Preliminaries}
\textbf{Basic Notation.} Let $\St=\{\s_1,...,\s_m \}$ be a training set with $m$ samples drawn from the input distribution $\D$.
As an adversarial sample $\s'$ is slightly different from a clean sample $\s$, we let $\St'$ and $\D'$ be an adversarial set and distribution for a specific model, respectively, such that $||\s'-\s||\le \epsilon$ (using by default the $\ell_2$-norm).
We omit the label $y$ of sample $\s$, as it is clear from the context. 
Let $\W$, $\W_l$ be the weight matrix, the weight matrix of the $l$-th layer, respectively. 
The loss function $\Lc(\cdot)$ and learning function $f_\W(\cdot)$ are parameterized over $\W$.
We consider $f_\W(\cdot)$ as the $n$-layer neural networks with $h$ hidden units per layer and activation functions $\act(\cdot)$.
We can now express each $f_\W(\s)$ as $f_\W(\s) = \W_{n}\act(\W_{n-1} ...\act(\W_1 \s)...)$. Note that we omit bias for convenience.
At the $l$-th layer ($l=1,\dots,n$), the latent representation before and after the activation function is denoted as $\h_l = \W_l\ac_{l-1}$ and $\ac_l = \act_l(\h_l)$, respectively.
We use $||\W_l||_2$ to denote the spectral norm of $\W_l$, deﬁned as the largest singular value of $\W_l$, and $||\W_l||_F$, to denote the Frobenius norm of $\W_l$. 
We denote the Kronecker product by $\otimes$ and the Hadamard product by $\odot$.

\textbf{Margin Loss.} For any margin $\gamma>0$, we deﬁne the expected margin loss as
\begin{equation}\nonumber
    \Lc_{\gamma,\D}(f_\W)=\Pro_{\s\sim 
    \D}\Big[ f_\W(\s)[y]\le \gamma+\max\limits_{j\ne y} f_\W(\s)[j] \Big],
\end{equation}
and let $\Lc_{\gamma,\St}(f_\W)$ be the empirical margin loss. Note that, setting $\gamma=0$ corresponds to the normal theoretical classification loss or empirical classification loss, which will be written as $\Lc_\D(f_\W)$ or $\Lc_\St(f_\W)$.

\section{Adversarial Generalization Bound}
\label{sec:bound}
Adversarial training 
is supposed to
improve the robustness against adversarial attacks not just on training samples, but also on unseen samples. Thus, the generalization performance of adversarial training (namely adversarial generalization, also known as robust generalization) may differ from that of non-adversarial natural training. 

For natural training,
PAC-Bayes \cite{mcallester1999pac,dziugaite2017computing} provides an upper bound on the generalization error with respect to the Kullback-Leibler divergence ($\KL$) between the posterior distribution $Q$ and the prior distribution $P$ of weights.
Let $f_\W$ be any predictor learned from the training data and parametrized by $\W$.
We consider the distribution $Q$ in the form of $f_{\W+\U}$
over learned weights $\W$, and denote by $\ul=\vecc(\U)$ (vectorization of $\U$) the multivariate random variable whose distribution may also depend on the training data.
Then, for any $\delta,\gamma>0$,  the following bound holds for the margin loss of any $f_\W$ with probability $1-\delta$ \cite{neyshabur2017pac,neyshabur2017exploring},
\begin{equation}
\label{eq:1}
\begin{aligned}
    \Lc_{\D}(f_\W) &\le  \Lc_{\gamma,\St}(f_\W)\\
    &+4 \sqrt{\frac{\KL(Q_{\vecc(\W)+\ul}||P)+\ln \frac{6m}{\delta}}{m-1}}.
\end{aligned}
\end{equation}
In the above expression, the $\KL$ term is evaluated, for a ﬁxed $\W$, with respect to the only random variable $\ul$. That is, the distribution of $\vecc(\W) + \ul$ 
can be obtained from
$\ul$ 
with the mean 
shifted by $\vecc(\W)$ and the same covariance matrix (and therefore the same
weight correlation matrix).

Notably, such a bound is so general that
the inequality holds for 
any
potential prior $P$ and posterior $Q$. Thus, 
specifying particular priors and posteriors 
does not violate the bound but only affects the tightness.
Furthermore, \cite{mcallester1999pac,neyshabur2017exploring} present a general framework to construct the posterior for a wide variety of models, including deterministic ones, to calculate the PAC-Bayesian bound. 

Given a learning setting, 
it is a common practice in the literature, \emph{e.g.,}
\cite{farnia2018generalizable,neyshabur2017pac,neyshabur2017exploring}, to \emph{assume that the prior $P$ to be a spherical Gaussian $\mathcal{N}(\vec{0},\sigma^2 \mathbf{I})$, and that the random variable $\ul$ also follows $\mathcal{N}(\vec{0},\sigma^2 \mathbf{I})$ (there is a slight difference between \cite{farnia2018generalizable} and \cite{neyshabur2017pac}, \cref{appendix:a}).} 
Under such assumption
in the adversarial setting
with
attack methods (\emph{e.g.,} FGM\cite{goodfellow2014explaining}, PGM\cite{kurakin2016adversarial}, WRM\cite{sinha2020certifying}), by letting the clean input domain be norm-bounded as $||\s||\le B$, $\forall \s \in \D$, where $B>0$ is a constant perturbation budget, 
and for any $\gamma,\delta>0$, any $\ul$ \emph{s.t.} $\Pro_\U(\max_{\s}||f_{\W+\U}(\s)-f_\W(\s)||<\frac{\gamma}{4})\ge\frac{1}{2}$~\cite{farnia2018generalizable,neyshabur2017pac}, \cite{farnia2018generalizable} gives the following margin-based PAC-Bayesian adversarial generalization bound,
\begin{equation}
\label{eq:2}
\begin{aligned}
&\Lc_{\D'}(f_\W) \le  \Lc_{\gamma,\St'}(f_\W)\\
& \quad +\mathcal{O} \Bigg( \sqrt{ \frac{(B+\epsilon)^2n^2h\ln(nh)\Phi^{\adv}+\ln\frac{m}{\delta}}{\gamma^2m}} \Bigg),
\end{aligned}
\end{equation}
where $\Phi^{\adv}$ depends on the attack method (and we omit the coefficient for $\ln\frac{m}{\delta}$ in \cref{eq:2}). \emph{E.g.,} for an FGM attack, let $\kappa\le ||\nabla_{\s''}\Lc(f_\W(\s''))||$ hold for every $\s''\in\{ \D \cup \D' \}$ with constant $\kappa>0$, we have 
\begin{equation}
\label{eq:3}
\begin{aligned}
    \Phi^{\adv} =&\prod_{l=1}^n ||\W_l||_2^2\Big\{ 1+\frac{\epsilon}{\kappa}(\prod_{l=1}^n||\W_l||_2)\\
    &\;\cdot \sum_{l=1}^n\prod_{j=1}^l ||\W_j||_2 \Big\}^2 \sum_{l=1}^n \frac{||\W_l||^2_F}{||\W_l||^2_2}.
\end{aligned}
\end{equation}
Details of other $\Phi^{\adv}$ are given in \cref{appendix:a}, and a proof is given in \cite{farnia2018generalizable}.
Note that we simplify \cref{eq:3} to $\Phi^{\adv}$, because our new terms of second-order statistics of weights in our bound are unrelated with it.

\subsection{Second-Order Statistics in Adversarial Bound}

The assumption of a spherical Gaussian distributed $\ul$ has greatly simplified the theoretical derivation of the bound in \cite{farnia2018generalizable,neyshabur2017pac,neyshabur2017exploring}.
Based on the same assumption,  \cite{farnia2018generalizable} develops an adversarial PAC-Bayesian bound by considering the impact of attack methods.
However, given the complexity of neural networks, this assumption is unrealistic. 


In this work, we relax this assumption by letting $\ul$ be a non-spherical Gaussian with the correlation matrix $\R$, where $\R \neq \mathbf{I}$ in general, and consider the impact of $\R$ on the above adversarial bound. 
Specifically, we assume
\emph{the 
correlations of weights from the same layer 
are not 0, whereas those from different layers are 0.}
Therefore, we develop the adversarial bound with the consideration of second-order statistics of weights. Before turning to the theoretical part, we first give the definition of $\R$.

\begin{mydef}
\label{def}
Given the clean sample $\s$ and the adversarial sample $\s'$, let $\ul_\s$ and $\ul_{\s'}$ be Gaussian distributed random vectors with each element being identically distributed as $\mathcal{N}(0,\sigma^2)$ but not independent one another.
Then over the entire datasets $\St$ and $\St'$,
$\ul_\St \triangleq \E_\s(\ul_\s)$ and $\ul_{\St'}\triangleq\E_{\s'}(\ul_{\s'})$
obey multivariate Gaussian mixture distributions, respectively, with corresponding correlation matrices as follows:
\begin{align}
\R_{\St} &\triangleq \frac{1}{\sigma^2} \E_\s[\E_{\ul}(\ul_\s \ul_\s^\T)],\\
\R_{\St'} &\triangleq \frac{1}{\sigma^2}\E_{\s'}[\E_{\ul}(\ul_{\s'} \ul_{\s'}^\T)].
\end{align}
\end{mydef}

To get a more general adversarial bound, it is reasonable to consider that the true correlation matrix $\R$ under adversarial training is over both $\St$ and $\St'$, rather than only over $\St'$, as most adversarial training methods may utilize both clean data and adversarial data \cite{engstrom2018evaluating,kannan2018adversarial,ding2018mma,wang2019improving,zhang2019theoretically,Zheng_Chen_Ren_2019,SJ2017,NEURIPS2020_5de8a360,zhang2020does,lee2020adversarial}. 
Therefore, we assume that the true $\ul$ of an adversarial trained model is a combination of two random variables, $\ul = q\ul_{\St}+(1-q)\ul_{\St'}$ and $\R = q\R_{\St}+(1-q)\R_{\St'}$, where $q\in[0,1]$ is an unknown parameter (as we are not clear how much clean data or adversarial data affects the model). 

Denote by $\R_{l,\St}$ and $\R_{l,\St'}$ the weight correlation matrices of $l$-th layer for clean and adversarial datasets, respectively, following Definition \ref{def}.
In the following, for notational convenience, we let
\begin{equation}
\begin{aligned}
\Lambda_{l,\max}&=\max\big(\lambda_{\max}(\R_{l,\St}),\lambda_{\max}(\R_{l,\St'})\big),\\
\Lambda_{l,\min}&=\min\big(\lambda_{\min}(\R_{l,\St}),\lambda_{\min}(\R_{l,\St'})\big),
\end{aligned}
\end{equation}
where $\lambda_{\max}(\cdot)$ and $\lambda_{\min}(\cdot)$ are the largest and the smallest singular value of the matrix, respectively. Note that 
$\R_{l,\St}$ and $\R_{l,\St'}$ are
symmetric positive semi-definite matrices, thus their eigenvalues and singular values coincide.


Next, by relaxing the spherical Gaussian assumption in \cref{eq:2} to consider the  non-spherical Gaussian distributed $\ul$ with the correlation matrix $\R$, we have the following lemma, as non-spherical Gaussian $\ul$ makes an obvious difference during the derivation of the $\KL$ term in \cref{eq:1} (\cref{appendix:b}).   
\begin{mylem}
\label{lemma:3.2}
Let the posteriori $Q$ be over the predictors of the form $f_{\W+\U}$, where $\ul$ is a non-spherical Gaussian with the correlation matrix $\R$. We can get
\begin{small}
\begin{equation}\nonumber
\begin{aligned}
\Lc_{\D'}(f_\W) &\le  \Lc_{\gamma,\St'}(f_\W)+\mathcal{O} \Bigg( \bigg(\frac{-\sum_l \ln \det \R_l+\ln\frac{m}{\delta}}{\gamma^2m}\\
&+ \frac{\Psi^{\adv}\big(\sum_l\big(c_1||\R'_l||_2^{\frac{1}{2}}+c_2||\R''_l||_2^{\frac{1}{2}}\big)\big)^{2}}{\gamma^2m} \bigg)^{\frac{1}{2}} \Bigg),
\end{aligned}
\end{equation}
\end{small}
where $c_1,c_2>0$ are universal constants and 
\begin{small}
\begin{equation}\nonumber
\begin{aligned}
    &\Psi^{\adv}=(B+\epsilon)^2\Phi^{\adv},\\
    &\R'_l=\E(\U_l^\T \U_l)/\sigma_l^2\\
    &=(\vec{I}^{h\times h}\otimes \vec{1}^{1\times h}) \big(\R_l\odot(\vec{1}^{h\times h}\otimes \vec{I}^{h\times h})\big)(\vec{I}^{h\times h}\otimes \vec{1}^{1\times h})^\T,\\
    &\R''_l=\E(\U_l \U_l^\T)/\sigma_l^2\\
    &=(\vec{I}^{h\times h}\otimes \vec{1}^{1\times h}) \big(\R_l\odot(\vec{I}^{h\times h}\otimes \vec{1}^{h\times h})\big)(\vec{I}^{h\times h}\otimes \vec{1}^{1\times h})^\T.
\end{aligned}
\end{equation}
\end{small} 
\end{mylem} We defer the \emph{proof} to \cref{appendix:b}. As mentioned in Def. \ref{def}, $\R_l$ is a combination of $\R_{l,\St}$ and $\R_{l,\St'}$ with an unknown coefficient $q$. We can use the following two lemmas to refine the bound in Lem. \ref{lemma:3.2} through the terms of  $\R_{l,\St}$ and $\R_{l,\St'}$.

\begin{mylem}
\vspace{-2mm}
\label{lemma:3.3}
$||\R'_l||_2^{\frac{1}{2}}$ and $||\R''_l||_2^{\frac{1}{2}}$ can be upper bounded by $\Lambda'_{l,\max}$ and $\Lambda''_{l,\max}$, i.e.,
\begin{equation}
\begin{aligned}
&||\R'_l||_2^{\frac{1}{2}} \le \Lambda'_{l,\max}, \; ||\R''_l||_2^{\frac{1}{2}} \le \Lambda''_{l,\max},\\
&\Lambda'_{l,\max}=\max\big(||\R'_{l,\St}||_2^{\frac{1}{2}},||\R'_{l,\St'}||_2^{\frac{1}{2}}\big),\\
&\Lambda''_{l,\max}=\max\big(||\R''_{l,\St}||_2^{\frac{1}{2}},||\R''_{l,\St'}||_2^{\frac{1}{2}}\big).
\end{aligned}
\end{equation}
\end{mylem}
\begin{custompro}{3.3}
\vspace{-2mm}
According to  \cite{knutson2001honeycombs}, we have 
\begin{small}
\begin{equation}\nonumber
\begin{aligned}
\lambda_{\max}(\R'_l)&\le q\lambda_{\max}(\R'_{l,\St})+(1-q)\lambda_{\max}(\R'_{l,\St'})\le (\Lambda'_{l,\max})^2,
\end{aligned}
\end{equation}
\end{small}and similarly $\lambda_{\max}(\R''_l)\le (\Lambda''_{l,\max})^2$.
\end{custompro}

\begin{mylem}
\vspace{-2mm}
\label{lemma:3.4}
The determinant of $\R_l$ can be lower bounded by the term of $\Lambda_{l,\min}$ and $\Lambda_{l,\max}$, i.e.,
\begin{equation}
    \det \R_l \ge \Lambda_{l,\min}^{k_l} \Lambda_{l,\max}^{\hh-k_l},
\end{equation}
where $k_l=(\hh\Lambda_{l,\max}-\hh)/(\Lambda_{l,\max}-\Lambda_{l,\min})$.
\end{mylem}
\begin{custompro}{3.4}
\vspace{-2mm}
For any vector $\x$, we have 
\begin{equation}
\begin{aligned}
    \langle \x, \R_l\x \rangle&=\langle \x, (q\R_{l,\St}+(1-q)\R_{l,\St'})\x \rangle\\
    &\ge (q\lambda_{\min}(\R_{l,\St})+(1-q)\lambda_{\min}(\R_{l,\St'}))||\x||^2\\
    &\ge \Lambda_{l,\min}||\x||^2.
\end{aligned}
\end{equation}
Hence, we can get $\lambda_{\min}(\R_l)\ge\Lambda_{l,\min}$. According to  \cite{knutson2001honeycombs}, we have $\lambda_{\max}(\R_l)\le \Lambda_{l,\max}$.
Finally, with the determinant lower bound in \cite{kalantari2001determinantal}, we can get Lem.~\ref{lemma:3.4} directly.
\end{custompro}
\vspace{-2mm}
Lems. ~\ref{lemma:3.2}, \ref{lemma:3.3} and \ref{lemma:3.4} lead to the following corollary.
\begin{mycor} 
\label{corollary}
Let $\ul$ be a non-spherical Gaussian with the correlation matrix $\R$ over $\St$ and $\St'$. Then we get 
\begin{footnotesize}
\begin{equation}
\begin{aligned}
    \Lc_{\D'}(f_\W) &\le \Lc_{\St',\gamma}(f_\W)+\mathcal{O}\Bigg( \bigg(\frac{ \Psi^{\adv}\big(\sum_l(c_1\Lambda'_{l,\max}+c_2\Lambda''_{l,\max})\big)^2}{\gamma^2 m}\\
    &\quad+\frac{+\ln\frac{m}{\delta}-\sum_l\ln (\Lambda_{l,\min}^{k_l} \Lambda_{l,\max}^{\hh-k_l})}{\gamma^2 m}\bigg)^{\frac{1}{2}} \Bigg).
\end{aligned}
\end{equation}
\end{footnotesize}
\end{mycor}
\begin{myrem}
\vspace{-2mm}
Cor.~\ref{corollary} demonstrates an updated adversarial generalization bound with consideration of non-spherical Gaussian $\ul$ over clean and adversarial data. It also indicates that, assume other coefficients are constant, minimizing $\Lambda'_{l,\max}$, $\Lambda''_{l,\max}$ and maximizing $\Lambda_{l,\min}^{k_l} \Lambda_{l,\max}^{\hh-k_l}$ can effectively tighten the adversarial generalization bound.
\end{myrem}

\section{Estimation and Optimization}
\label{sec:estimate}
According to Cor.~\ref{corollary}, we need to monitor and control the weight correlation matrix and some of its norms -- such as the singular value, the spectral norm, and the determinant -- during training. To this end, we need to be able to efficiently estimate weight correlation matrix and have a corresponding effective optimization scheme for training. 
\subsection{Estimation of the Weight Correlation Matrix}
We employ two different methods to estimate the weight correlation matrix and, through an inter-comparison between their estimations, to ensure that our empirical conclusions are not compromised by the estimation errors. One is a sampling method, and the other is Laplace approximation of neural networks~\cite{botev2017practical,ritter2018scalable}.
Note that, though we only use Laplace approximation to optimize the second-order statistics terms (S$^2$O) during training 
due to time complexity of sampling (\cref{sec:optimize}), our empirical results in \cref{sec:5.1} indicate that S$^2$O is 
applicable with both methods. 

\vspace{-3mm}
\paragraph{\textbf{Sampling method}} obtains a
set of weight samples $(\W+\eta)$ 
by a sharpness-like method~\cite{keskar2016large,jiang2019fantastic} \emph{s.t.}  $|\Lc(f_{\W+\eta})-\Lc(f_{\W})|\le \epsilon'$ (\emph{e.g.,} $\epsilon'=0.05$ for CIFAR-10 and $\epsilon'=0.1$ for CIFAR-100), where $\vecc(\eta)$ is a $\vec{0}$ mean Gaussian noise. 
These samples are then used 
to estimate the correlation matrix of $\ul_\St$ and $\ul_{\St'}$.
More details are given in \cref{appendix:c}.  

\vspace{-3mm}
\paragraph{\textbf{Laplace approximation}} 
is 
an estimation method 
widely used in Bayesian framework to approximate posterior densities or posterior moments \cite{ritter2018scalable,tierney1986accurate,rue2009approximate}. Technically, it 
approximates the posterior (\emph{e.g.,} $\vecc(\W)+\ul$) by a Gaussian distribution with the second-order Taylor expansion of the $\ln$ posterior around its MAP estimate. Specifically, given weights for layer $l$ with an MAP estimate $\W_l^*$ on $\St$ (we omit the estimation on $\St'$ as it is the same with $\St$), we have
\begin{small}
\begin{equation}
\begin{aligned} \label{eq:laplace-approx}
   &\ln p(\vecc(\W_l)+\ul_l|\St) \approx \ln p\Big(\vecc(\W_l^*)|\St\Big)\\
   &-\frac{1}{2}\Big(\vecc(\W_l-\W_l^*)+\ul_l\Big)^\T {\E_{\s}[\HH_{l}]} \Big(\vecc(\W_l-\W_l^*)+\ul_l\Big),
\end{aligned}
\end{equation}
\end{small}where $\E_{\s}[\HH_{l}]$ is the expectation of the Hessian matrix over input data sample $\s$, 
and the Hessian matrix $\HH_{l}$ 
is given by
$\HH_{l}=\frac{\partial^2 \Lc(f_{\W}(\s))}{\partial \vecc(\W_l)\partial \vecc(\W_l)}$.

It should be noted that, in \cref{eq:laplace-approx}, the first-order Taylor polynomial has been dropped because the gradient around the MAP estimate $\W_l^*$ is zero.
Then, taking a closer look at \cref{eq:laplace-approx}, we find that its second line is exactly the logarithm of the probability density function of a Gaussian distributed multivariate random variable with mean $\W^*_l$ and covariance ${\Sigma_{l}}=\E_{\s}[\HH_{l}]^{-1}$, \emph{i.e.},
$\vecc(\W_l)+\ul_l \sim \mathcal{N}(\vecc(\W^{*}_l),\Sigma_{l})$,
where $\Sigma_{l}$  can be viewed as the covariance matrix of $\ul_l$ and learned weights $\W_l$ can be seen as the MAP estimate $\W_l^*$.

Laplace approximation indicates that 
efficiently estimating $\Sigma_{l}$ is achievable through the inverse of the Hessian matrix, because ${\Sigma^{-1}_{l}}=\E_{\s}[\HH_{l}]$. \emph{Note that we omit $\Sigma^{-1}_{l,\St'} = \E_{s'}[\HH_{l}]$ as it is similar to $\Sigma^{-1}_{l,\St} = \E_{s}[\HH_{l}]$.}
Moreover, \cite{botev2017practical,ritter2018scalable} developed a Kronecker factored Laplace approximation based on insights from second-order optimization of neural networks. That is, in contrast to the classical second-order methods~\cite{battiti1992first,shepherd2012second} with high computational costs for deep neural networks, they suggest that Hessian matrices of $l$-th layer can be Kronecker factored, i.e., 
\vspace{-2mm}
\begin{equation} \label{eq:hessian-decompose}
\begin{aligned}
    \HH_{l}=\underbrace{\ac_{l-1}\ac_{l-1}^\T}_{\mathcal{A}_{l-1}} \otimes \underbrace{\frac{\partial^2 \ell(f_{\W}(\s))}{\partial \h_{l}\partial \h_{l}}}_{\mathcal{H}_{l}}=\mathcal{A}_{l-1} \otimes \mathcal{H}_{l},
\end{aligned}
\vspace{-2mm}
\end{equation}
where $\mathcal{A}_{l-1} \in \mathbb{R}^{h \times h}$ is the covariance of
the post-activation of the previous layer, and $\mathcal{H}_{l} \in \mathbb{R}^{h \times h}$ is the Hessian matrix of the loss with respect to the pre-activation of the current layer, and $h$ is the number of neurons for each layer. With the assumption that $\mathcal{A}_{l-1}$ and $\mathcal{H}_{l}$ are independent~\cite{botev2017practical,ritter2018scalable}, we can approximate $\E_{\s}[\HH_{l}]$ with
\begin{equation}
\label{eq11}
\begin{aligned}
    \E_{\s}[\HH_{l}] = \E_{\s}[\mathcal{A}_{l-1} \otimes \mathcal{H}_{l}]\approx \E_{\s}[\mathcal{A}_{l-1}] \otimes \E_{\s}[\mathcal{H}_{l}].
\end{aligned}
\end{equation}

\begin{figure*}[t!]
\includegraphics[width=1
\textwidth]{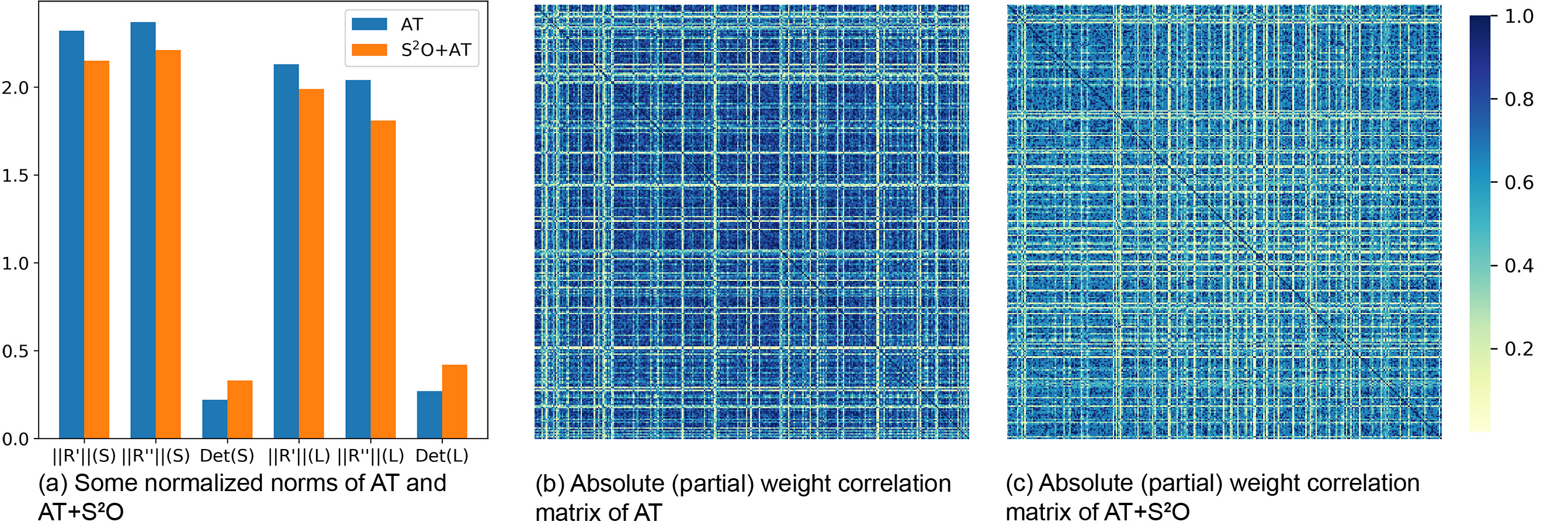}
\centering
\vspace{-5mm}
\caption{We train PreAct ResNet18 with AT and AT+S$^2$O, and show the results of partial weights. \textbf{(a)} shows the normalized spectral norm of $\R'_{\St'}$, ${\R''}_{\St'}$, and the determinant of $\R_{\St'}$, with sampling estimation (S) and Laplace approximation (L) respectively. \textbf{(b)} and \textbf{(c)} demonstrate the absolute correlation matrix of partial weights, for AT and AT+S$^2$O respectively.}   
\vspace{-2mm}
\label{fig:2}
\end{figure*}

\subsection{A Novel Optimization Scheme S$^2$O for Training}
\label{sec:optimize}
The adversarial training of a neural network is seen as a process of optimizing over an adversarial objective function $J_{\adv}$. To tighten the adversarial bound in Cor. \ref{corollary}, we add the second-order statistics penalty terms $\Lambda'_{l,\max}$, $\Lambda''_{l,\max}$ and $-\ln\Lambda_{l,\min}^{k_l} \Lambda_{l,\max}^{\hh-k_l}$ to the objective function $J_{\adv}$, and denote the new objective function as $\tilde{J}_{\adv}$. To reduce the complexity, we approximate $\nabla_{\w_l}(\Lambda'_{l,\max}+\Lambda''_{l,\max}-\ln\Lambda_{l,\min}^{k_l} \Lambda_{l,\max}^{\hh-k_l})$
through
\begin{equation}
    \nabla_{\w_l}(\g(\R_{l}))=\frac{\partial\big(||\R_{l,\St}||_F^2+||\R_{l,\St'}||_F^2\big)}{\partial \vecc(\W_l)}.
\label{eq:R}
\end{equation}
Although it is impractical to find the exact relation between $\R_l$ and above penalty terms (\emph{e.g.}, perfect positive or negative),  they are clearly related. 
In particular, when $\R_{l,\St}$ and $\R_{l,{\St'}}$ have the same off-diagonal elements as $r_\s$ and $r_{\s'}$, respectively, and $r_\s r_{\s'}\ge 0$, we have the following lemmas.
\begin{mylem}
\label{lem:4.1}
Decreasing $||\R_{l,\St}||_F^2$ and $||\R_{l,\St'}||_F^2$ leads to a decline in $|r_{\s}|$ and $|r_{\s'}|$, and further causes reduced  $\Lambda'_{l,\max}$ and $\Lambda''_{l,\max}$.    
\end{mylem}
\begin{mylem}
\label{lem:4.2}
Decreasing $||\R_{l,\St}||_F^2$ and $||\R_{l,\St'}||_F^2$ leads to an increase in $\Lambda_{l,\min}^{k_l} \Lambda_{l,\max}^{\hh-k_l}$.
\end{mylem}
Proofs are given in \cref{appendix:d}. We also provide more general case simulations of the relationship between $||\R_{l,\St}||_F^2$, $||\R_{l,\St'}||_F^2$ and the above penalty terms in \cref{appendix:e}.

\begin{myrem}
Lems.~\ref{lem:4.1}, \ref{lem:4.2} and simulations in \cref{appendix:e} indicate that we can 
decrease $\Lambda'_{l,\max}$, $\Lambda''_{l,\max}$ and increase $\Lambda_{l,\min}^{k_l} \Lambda_{l,\max}^{\hh-k_l}$ 
by decreasing $||\R_{l,\St}||_F^2$ and $||\R_{l,\St'}||_F^2$.   
\end{myrem}

It is noted that a direct optimization over \cref{eq:R} is also computationally prohibitive.
Fortunately, Laplace approximation can greatly reduce the complexity of \cref{eq:R}.
Specifically, according to \cref{eq11} and ${\Sigma_l}=\E[\HH_{l}]^{-1}$, the following term
\begin{equation}
    \nabla_{\w_{l-1}}(\g(\A_{l-1}))=\frac{\partial\big(||\A_{l-1,\St}||_F^2+||\A_{l-1,\St'}||_F^2\big)}{\partial \vecc(\W_{l-1})},
\label{eq:13}
\end{equation}
can be used to approximate $\nabla_{\w_l}(\g(\R_{l}))$, where $\A_{l-1,\St}$ is the normalization of $\E_{\s}[\mathcal{A}_{l-1}]^{-1}$, \emph{i.e.,} $\forall \; 0<i,j\le h$ and for  $i,j \in \N$,
\begin{equation}
    (\A_{l-1,\St})_{[ij]}=\frac{(\E_{\s}[\mathcal{A}_{l-1}]^{-1})_{[ij]}}{\sqrt{(\E_{\s}[\mathcal{A}_{l-1}]^{-1})_{[ii]}(\E_{\s}[\mathcal{A}_{l-1}]^{-1})_{[jj]}}}.
\end{equation}

Finally, we add the regularizer $\g(\A)$ to the adversarial training objective function $J_{\adv}$, obtaining the new objective function $\tilde{J}_{\adv}$ with
\begin{equation}
    \nabla_{\w}\tilde{J}_{\adv}=\nabla_{\w}J_{\adv}+\alpha\nabla_{\w}\g(\A).
\end{equation}
Here, $\alpha \in [0,\infty)$ is a hyper-parameter to balance the relative contributions of the second-order statistics penalty term $\g(\A)$ and the original objective function $J_{\adv}$ (\cref{appendix:f}). 

\begin{table*}[t]
\centering
\caption{Adversarial training across data sets on PreAct ResNet18 (\%).}
\label{tab:1}
\vspace{-2mm}
\scalebox{0.85}{
\begin{tabular}{clccccccccccc}
\toprule[1.5pt]
Threat &\multirow{2}{*}{Method} & \multicolumn{3}{c}{CIFAR-10} && \multicolumn{3}{c}{CIFAR-100} && \multicolumn{3}{c}{SVHN} \\
\cline{3-5} \cline{7-9} \cline{11-13}
Model&& Clean & PGD-20 & Time/epoch & & Clean & PGD-20 & Time/epoch & & Clean & PGD-20 & Time/epoch         
\\ \hline
\multirow{2}{*}{$\ell_\infty $} &AT & 82.41 & 52.77 & 309s & & 58.02 & 28.02 & 307s & & 93.17 & 60.91 & 509s \\
&AT+S$^2$O & \textbf{83.65} & \textbf{55.11} & 368s & & 58.45 & \textbf{30.58} & 371s & & 93.39 &\textbf{64.83} & 595s  \\ 
\hline
\multirow{2}{*}{$\ell_2 $} &AT & 88.83 & 68.83 & 292s & & 64.21 & 42.20 & 290s & & 94.02 & 66.76 & 477s \\
&AT+S$^2$O & \textbf{89.57} & \textbf{69.42} & 364s & & \textbf{65.32} & \textbf{44.07} & 366s & & \textbf{94.93} & \textbf{76.19} & 586s \\ 
\toprule[1.5pt]
\end{tabular}
}
\label{table:1}
\vspace{-3mm}
\end{table*}
\begin{table*}[t!]
\centering
\caption{TRADES ($1/\lambda=6$) and AWP on CIFAR-10 with $\ell_\infty$ threat model (\%).}
\label{tab:2}
\vspace{-2mm}
\scalebox{0.82}{
\begin{tabular}{lccccccccccccc}
\toprule[1.5pt]
\multirow{2}{*}{Method} & \multicolumn{6}{c}{ResNet18} && \multicolumn{6}{c}{WideResNet} \\
\cline{2-7} \cline{9-14}
& Clean & FGSM & PGD-20 & PGD-100 & CW-20 & AA & & Clean & FGSM & PGD-20 & PGD-100 & CW-20 & AA         \\ \hline 

TRADES & 82.89 & 58.72 & 53.81 & 53.69 & 51.83 & 48.6 & & 83.98 & 61.08 & 56.82 & 56.53 & 54.54 & 52.7 \\
TRADES+AWP  & 82.30 & 59.48  & 56.18 & 55.90 & 53.12 & 51.7 & & 84.99 & 63.11 & 59.67 & 59.42 & 57.41 & \textbf{56.2} \\
TRADES+S$^2$O & \textbf{84.15} & 60.19 & 55.20 & 54.73 & 52.47 & 49.5 & & 85.67 & 62.73 & 58.34 & 57.69 & 55.36 & 54.1 \\
TRADES+AWP+S$^2$O & 83.79 & \textbf{60.27} & \textbf{57.29} & \textbf{56.51} & \textbf{53.84} & \textbf{52.4} & & \textbf{86.01} & \textbf{64.16} & \textbf{61.12} & \textbf{60.46} & \textbf{57.93} & 55.9 \\

\toprule[1.5pt]
\end{tabular}
}
\label{table:PGD-training}
\vspace{-5mm}
\end{table*}

\section{Empirical Results}
\label{sec:experiments}
In this section, we first provide a comprehensive understanding of our S$^2$O training method, and then evaluate its robustness on benchmark data sets against various white-box and black-box attacks. 

\textbf{Experimental Setup.} We train PreAct ResNet-18 \cite{he2016deep} for $\ell_\infty$ and $\ell_2$ threat models on CIFAR-10/100 \cite{krizhevsky2009learning} and SVHN \cite{netzer2011reading} (\cref{tab:1}). 
In addition, we also train  WideResNet-34-10~\cite{zagoruyko2016wide} for CIFAR-10 with an $\ell_\infty$ threat model (\cref{tab:2,tab:3}).
We follow the settings of \cite{rice2020overfitting}: 
for the $\ell_\infty$ threat model, $\epsilon=8/255$ and step size $2/255$;
for the $\ell_2$ threat model, $\epsilon=128/255$ and step size $15/255$ for all data sets.
In all experiments, the training/test attacks are PGD-10/(PGD-20 and others) respectively.
All models (except SVHN) are trained for $200$ epochs using SGD with momentum $0.9$, batch size $128$, weight decay $5\times 10^{-4}$, and an initial learning rate of $0.1$ that is divided by $10$ at the 100th and 150th epochs.
For
SVHN, we use the same parameters except for setting the starting
learning rate to $0.01$.
Simple data augmentations, such as $32\times 32$ random crop with $4$-pixel padding and random horizontal flip, are applied. 
We implement each PreAct ResNet18 on single GTX 1080 Ti and each WideResnet on single NVIDIA A100. 

\textbf{White-box attack.} We conduct white-box attacks, including FGSM~\cite{goodfellow2014explaining}, PGD-20/100 \cite{madry2017towards}, and CW-20~\cite{carlini2017towards} (the $\ell_\infty$ version of CW loss optimized by PGD-20), on the models trained with baseline methods and our S$^2$O-enhanced variants. 

\textbf{Black-box attack.} Black-box 
attacks are created from the clean test data by attacking a surrogate model with an architecture that is either a copy of the defense model or a more complex model~\cite{papernot2017practical}. After constructing adversarial examples from each of the trained models, we apply these adversarial examples to the other models and evaluate the performances. The attacking methods we have used are FGSM and PGD-20.

\textbf{Auto Attack.} We 
consider Auto Attack (AA)~\cite{croce2020reliable}, a powerful and reliable attack, which 
attacks
through an ensemble of different parameter-free attacks that include three white-box attacks (APGD-CE~\cite{croce2020reliable}, APGD-DLR~\cite{croce2020reliable}, and FAB~\cite{croce2020minimally}) and a black-box attack (Square Attack~\cite{andriushchenko2020square}).

By default, we use the setting $\alpha=0.3$ for S$^2$O, with the exception of $\alpha=0.1$ for AT+S$^2$O (CIFAR-10) in \cref{tab:1}. 
Following \cite{zhang2019theoretically}, we set epsilon=0.031and step size=0.003 for PGD and CW evaluation.
And we use standard version auto attack evaluation.  

\begin{table}[t!]
\centering
\captionsetup{format=myformat}
\caption{AVMixup and MART on CIFAR-10 with $\ell_\infty$ threat model for WideResNet (\%).}
\label{tab:3}
\vspace{-2mm}
\scalebox{0.74}{
\begin{tabular}{lcccccc}
\toprule[1.5pt]
Method & Clean & FGSM & PGD-20 & PGD-100 & CW-20 & AA          \\ \hline 
AVMixup & 92.56 & 80.46 & 59.75 & \gaojie{49.51} & 54.53 & \textcolor{cyan}{39.7}  \\
AVMixup+S$^2$O & \textbf{93.72} & \textbf{84.57} & \textbf{60.43} & \gaojie{50.49} & \textbf{56.16} & \textcolor{cyan}{39.3} \\
\hline
MART & 83.51 & 61.53 & 58.31 & 57.55 & 54.33 & 51.2 \\
MART+S$^2$O & 83.91 & \textbf{62.56} & \textbf{59.29} & \textbf{58.33} & \textbf{55.14} & \textbf{54.1} \\
\toprule[1.5pt]
\end{tabular}
}
\vspace{-5mm}
\end{table}

\vspace{-2mm}
\subsection{Empirical Understanding of S$^2$O}
\vspace{-2mm}
\label{sec:5.1}
In this part, we explore how the second-order statistics of weights (\emph{e.g.,} weight correlation matrix)  change when we apply S$^2$O on it.
The results in \cref{fig:2} indicate that S$^2$O effectively decreases the spectral norm of $\R'_{\St'}$, ${\R''}_{\St'}$ and increases the determinant of $\R_{\St'}$. We also provide the results of clean data in \cref{appendix:e}.

\begin{table}[t!]
\centering
\captionsetup{format=myformat}
\caption{TRADES ($1/\lambda=6$) and AWP on CIFAR-100 with $\ell_\infty$ threat model for WideResNet (\%).}
\label{tab:4}
\vspace{-3mm}
\scalebox{0.74}{
\begin{tabular}{lccccc}
\toprule[1.5pt]

Method & Clean & FGSM & PGD-20 & CW-20 & AA \\ \hline 

TRADES & 60.38 & 35.01 & 32.11 & 28.93 & 26.9 \\
TRADES+LBGAT  & 60.43 & - & 35.50 & \textbf{31.50} & 29.3 \\
TRADES+AWP & 60.27 & 36.12 & 34.04 & 30.64 & 28.5 \\
TRADES+S$^2$O & 63.40 & 35.96 & 33.06 & 29.57 & 27.6  \\
TRADES+AWP+S$^2$O & \textbf{64.17} & \textbf{37.98} & \textbf{35.95} & 31.26 & \textbf{29.9} \\

\toprule[1.5pt]
\end{tabular}
}
\vspace{-3mm}
\end{table}
\begin{table}[t!]
\centering
\captionsetup{format=myformat}
\caption{VGG16 and MobileNetV2 on CIFAR-10 with $\ell_\infty$ threat model (\%).}
\label{tab:5}
\vspace{-3mm}
\scalebox{0.74}{
\begin{tabular}{lccccc}
\toprule[1.5pt]
Method & Clean & FGSM & PGD-20 & CW-20 & AA
\\ \hline
VGG16 AT & 81.63 & 53.23 & 49.21 & 48.01 & 43.1   \\
VGG16 AT+S$^2$O & \textbf{82.57} & \textbf{54.03} & \textbf{50.53} & 48.15 & \textbf{44.6}   \\
\hline
MNV2 AT & 81.97 & 55.52 & 50.76 & 49.53 & 44.9   \\
MNV2 AT+S$^2$O & \textbf{82.48} & \textbf{57.51} & \textbf{52.93} & 49.92 & \textbf{45.7}   \\
\toprule[1.5pt]
\end{tabular}}
\vspace{-5mm}
\end{table}

\subsection{Robustness Under White-Box Attacks and Auto Attack}
\label{sec:5.2}

\vspace{-1mm}
\paragraph{Applying \so\ on vanilla adversarial training (\cref{table:1}).}
We employ PreAct ResNet-18 to explore the power of our proposed S$^2$O method embedded with normal PGD-10 training (with $\ell_{\infty}$ and $\ell_2$ threat models), across a number of data sets including CIFAR-10, CIFAR-100, and SVHN.

\cref{table:1} suggests that S$^2$O-enhanced variants can improve both, the accuracy (over clean data) and the robust accuracy (over PGD-20 attack), across the three datasets. For example, the accuracy on PGD-20 of the AT+S$^2$O model is 2\%-3\% higher than that of the standard adversarial training model with an $\ell_{\infty}$ threat model on CIFAR-10. There is also an increase of 1\%-1.5\% in accuracy on clean data when compared to the standard adversarial training model. Generally, the improvement is very consistent across data sets and attacks.

\vspace{-4mm}
\paragraph{Applying \so\ on TRADES and AWP (\cref{tab:2}).} We 
employ PreAct ResNet-18 and WideResNet to explore the performance of our S$^2$O method when it works with two state-of-the-art methods, TRADES and TRADES+AWP, on the CIFAR-10 (under an $\ell_{\infty}$ threat model). 
The robustness of all defense models are tested against white-box FGSM, PGD-20, PGD-100, CW-20 attacks, and auto attack.

\cref{tab:2} shows that S$^2$O-enhanced variants perform consistently and significantly better than the existing ones (with only one exception).
While AWP improves over TRADES, S$^2$O  can further enhance it.
For example, the accuracy on PGD-20 of the S$^2$O-enhanced TRADES+AWP model is 1.45\% higher than that of the TRADES+AWP model on WideResNet, and the accuracy on PGD-20 of the S$^2$O-enhanced TRADES model is 1.39\% higher than that of the TRADES model on PreAct ResNet-18. For additional experiments on CIFAR-100 in \cref{tab:4}, compared with TRADES+LBGAT~\cite{cui2021learnable}, S$^2$O can also improve robustness under most attacks.

\vspace{-4mm}
\paragraph{Applying \so\ on AVMixup and MART (\cref{tab:3}).} We 
employ WideResNet to study the performance of our S$^2$O method when it works with some other state-of-the-art methods, such as AVMixup and MART, on the CIFAR-10 data set (under $\ell_{\infty}$ threat model). 
The robustness of all defense models is tested against white-box FGSM, PGD-20, PGD-100, CW-20 attacks, and auto attack.

\cref{tab:3} also shows that S$^2$O enhanced models 
perform better than normal AVMixup and MART under most attacks (and clean data).
Note that we mark the results of AVMixup with the color \gaojie{cyan} under PGD-100 attack and auto attack (AA), as AVMixup (including AVMixup+S$^2$O) is not a robust method across all attacks -- it performs not so well under PGD-100 and Auto attacks.

\vspace{-4mm}
\paragraph{Remark on our Baselines in \cref{tab:1,tab:2,tab:3,tab:4}.} We have checked that our baselines are close to, or slightly better than, the baselines in the recent paper \cite{wu2020adversarial}, with which we have very similar experimental settings. We omit
the standard deviations of 3 runs as they are very small ($<0.40\%$).

\vspace{-4mm}
\paragraph{Supplement.} We provide hyper-parameter ($\alpha$) sensitivity analysis in \cref{tab:6} with PreAct ResNet-18 and CIFAR-10.
We also apply S$^2$O to other structures (VGG16~\cite{simonyan2014very} and  MobileNetV2~\cite{sandler2018mobilenetv2}) with normal adversarial training; \cref{tab:5} shows that S$^2$O also works on these two structures.
In addition, we notice that SOAR~\cite{ma2020soar} can get 56.06\% accuracy on ResNet-10, CIFAR-10 under PGD-20 attack; it is interesting to combine S$^2$O with SOAR in the future work.  

\begin{table}[t]
\centering
\captionsetup{format=myformat}
\caption{Sensitivity analysis  on CIFAR-10 with $\ell_\infty$ threat model for ResNet18 (\%).}
\label{tab:6}
\vspace{-3mm}
\scalebox{0.74}{
\begin{tabular}{l|cc|ccc}
\toprule[1.5pt]
Method & Clean & AA & PGD-20$_{train}$ & PGD-20$_{test}$ & Gap 
\\ \hline
AT & 82.41 & 47.1 & 62.33 & 52.77 & 9.56  \\
AT+S$^2$O (0.05) & 83.22 & \textbf{48.5} & 61.99 & 53.82 & 8.17  \\
AT+S$^2$O (0.1) & \textbf{83.65} & 48.3 & 61.50 & \textbf{55.11} & 6.39   \\
AT+S$^2$O (0.2) & 83.43 & 47.8 & 60.27 & 54.59 & 5.68  \\ 
AT+S$^2$O (0.3) & 82.89 & 46.5 & 59.36 & 54.24 & 5.12   \\
AT+S$^2$O (0.4) & 82.54 & 46.7 & 58.41 & 52.92 & 5.49   \\ 
\toprule[1.5pt]
\end{tabular}}
\vspace{-3mm}
\end{table}
\begin{table}[t!]
\centering
\captionsetup{format=myformat}
\caption{Adversarial training across data sets on PreAct ResNet18 with black-box attacks under $\ell_\infty$ threat model (\%).}
\label{tab:7}
\vspace{-2mm}
\scalebox{0.74}{
\begin{tabular}{clcc}
\toprule[1.5pt]
Data & Method & FGSM & PGD-20 \\ \hline 
\multirow{2}{*}{CIFAR-10} & AT & 64.32 & 62.63  \\
& AT+S$^2$O & \textbf{65.63} & \textbf{63.87} \\
\multirow{2}{*}{CIFAR-100} & AT & 38.55 & 37.36  \\
& AT+S$^2$O & \textbf{39.68} & \textbf{38.60}  \\
\multirow{2}{*}{SVHN} & AT & 71.77 & 63.76  \\
& AT+S$^2$O & 72.20 & \textbf{64.31} \\
\toprule[1.5pt]
\end{tabular}
}
\vspace{-5mm}
\end{table}

\vspace{-2mm}
\subsection{Robustness Under Black-Box Attacks}
\vspace{-2mm}
\label{sec:5.3}
We also employ PreAct ResNet-18 to explore the power of our proposed S$^2$O method embedded with normal PGD-10 training under black-box attacks (with an $\ell_{\infty}$ threat model), across a number of data sets, including CIFAR-10/100, and SVHN. For the same data set, all black-box attacks are generated by the same adversarial training model. 

\cref{tab:7} suggests that S$^2$O enhanced models also can get some improvements under black box attacks.  

\vspace{-2mm}
\section{Related Work}
\vspace{-2mm}
Adversarial training updates the minimization objective of the  training scheme from the usual one to 
\vspace{-2mm}
\begin{equation}
    J_{\mathrm{adv}}=\mathop{\E}\limits_{\s\sim \D} \bigg[\max_{||\s'-\s||\le \epsilon} \ell(f_\W(\s'))\bigg],
\label{eq1}
\vspace{-1mm}
\end{equation}
where $\s'$ is an adversarial example causing the greatest loss within an $\epsilon$-ball centered at a clean example $\s$ with respect to a norm distance (by default $\ell_2$).
Adversarial training methods roughly fall within three categories, and in the following, we highlight four state-of-the-art 
methods, which are combined with our \so\ method in the experiments. 

The first category is to reduce \cref{eq1} to an equivalent, or approximate, expression, which includes measuring the distance between $\s$ and $\s'$.
For example, ALP ~\cite{engstrom2018evaluating,kannan2018adversarial} enforces the similarity between $f_\W(\s)$ and $f_\W(\s')$, the logits activations on unperturbed and adversarial versions of the same image $\s$. 
MMA ~\cite{ding2018mma} encourages every correctly classified instance $\s$ to leave sufficiently large margin, i.e., the distance to the boundary, by maximizing the size of the shortest successful perturbation. 
\textbf{MART} \cite{wang2019improving} observes the difference of misclassified and correctly classified examples in adversarial training, and suggests different loss functions for them. 
\textbf{TRADES} \cite{zhang2019theoretically} analyzes the robustness error and the clean error, and shows an upper bound and lower bound on the gap between robust error and clean error, which motivates adversarial training networks to optimize $J_{\mathrm{tr}}$ with
\vspace{-2mm}
\begin{small}
\begin{equation}
\label{eq_trades}
    \mathop{\E}\limits_{\s\sim \D}\Big[ \ell(f_\W(\s))+\max_{||\s'-\s||\le \epsilon}\KL\big(f_\W (\s)||f_\W (\s')\big)/\lambda\Big],
\vspace{-2mm}
\end{equation}
\end{small}where $\lambda$ is the hyper-parameter to control the trade-off between clean accuracy and robust accuracy. 
It considers the $\KL$-divergence of the activations of the output layer, i.e., $\KL(f_\W(\s)||f_\W(\s'))$, for every instance $\s$.
The measurement over $\s$ and $\s'$ can be extended to consider a \emph{local} distributional distance, i.e., the distance between the distributions within a norm ball of $\s$ and within a norm ball of $\s'$. 
For example, \cite{Zheng_Chen_Ren_2019} forces the similarity between local distributions of an image and its adversarial example, \cite{SJ2017} uses Wasserstein distance to measure the similarity of local distributions, and  \cite{dong2020adversarial,NEURIPS2020_5de8a360,dong2020benchmarking,mao2019metric,pang2020bag} optimize over distributions over a set of adversarial perturbations for a single image.   

The second category is to pre-process the generated adversarial examples before  training instead of directly using the adversarial examples generated by attack algorithms.
Notable examples include label smoothing \cite{szegedy2016rethinking,chen2020robust}, which, instead of considering the adversarial instances $(\s',y)$ for the ``hard'' label $y$, it considers $(\s',\tilde{y})$, where $\tilde{y}$ is a ``soft'' label represented as a weighted sum of the hard label and the uniform distribution. 
This idea is further exploited in  \cite{muller2019does}, which empirically studies how label smoothing works.
Based on these,  \textbf{AVMixup} \cite{zhang2020does,lee2020adversarial} defines a  virtual sample in the adversarial direction and extends the training distribution with soft labels via linear interpolation of the virtual sample and the clean sample.
Specifically, it optimizes 
\vspace{-2mm}
\begin{equation}
\label{eq_avm}
    J_{\mathrm{avm}}=\mathop{\E}\limits_{\s\sim \D} \Big[ \ell(f_\theta(\hat\s),\hat \y)\Big],
\vspace{-2mm}
\end{equation}
where $\hat \s=\beta\s+(1-\beta)\gamma(\s'-\s)$, $\hat \y=\beta\phi(\y,\lambda_1)+(1-\beta)\phi(\y,\lambda_2)$, $\beta$ is drawn from the Beta distribution for each single $\s_i$, $\gamma$ is the hyper-parameter to control the scale of adversarial virtual vector, $\y$ is the one-hot vector of $y$,  $\phi(\cdot)$ is the label smoothing function~\cite{szegedy2016rethinking}, and $\lambda_1$ and $\lambda_2$ are hyper-parameters to control the smoothing degree. 
Other than label smoothing, \cite{NEURIPS2019_d8700cbd} generates adversarial examples by perturbing the local neighborhoods in an unsupervised fashion.

These two categories follow the min-max formalism and only adapt its components. 
\textbf{AWP} \cite{wu2020adversarial} 
adapts the inner maximization 
to take one additional maximization to find a weight perturbation based on the generated adversarial examples. The outer minimization is then based on the perturbed weights~\cite{devries2017improved} to minimize the loss induced by the adversarial examples.
Specifically, it is to
optimize the double-perturbation adversarial training problem
\begin{equation}
\label{eq_awp}
    J_{\mathrm{awp}}=\max_{\vec{V}\in\V}\mathop{\E}\limits_{\s\sim \D}\max_{||\s'-\s||\le \epsilon}\ell( f_{\W+\vec{V}}(\s') ),
\end{equation}
where $\V$ is a feasible region for the parameter perturbation~$\vec{V}$. 


Another thread of related work is on  the PAC-Bayesian framework, a well-known theoretical tool to bound the generalization error of machine learning models~\cite{mcallester1999pac,langford2001bounds,maurer2004note,langford2002not,langford2003pac,germain2009pac,parrado2012pac,welling2011bayesian}. Recently, it is also widely developed in various aspects for both traditional machine learning models and deep neural networks~\cite{thiemann2017strongly,rivasplata2019pac,dziugaite2018data,dwork2015generalization,dwork2015preserving,blundell2015weight,alquier2016properties,letarte2019dichotomize,jin2020does,jin2022weight,perez2021tighter}.




\section{Conclusion}

This work addresses an oversight in the adversarial training literature by arguing that the second-order statistics of weights need to be systematically considered. Through theoretical study (updating the PAC-Bayesian framework), algorithmic development (efficient estimation of weight correlation matrix,  effective training with \so), and extensive experiments, we show that the consideration of second-order statistics of weights can improve the robustness and generalization over not only the vanilla adversarial training but also the state-of-the-art adversarial training methods. 

\vspace{-3mm}
\paragraph{Acknowledgment.} \includegraphics[height=8pt]{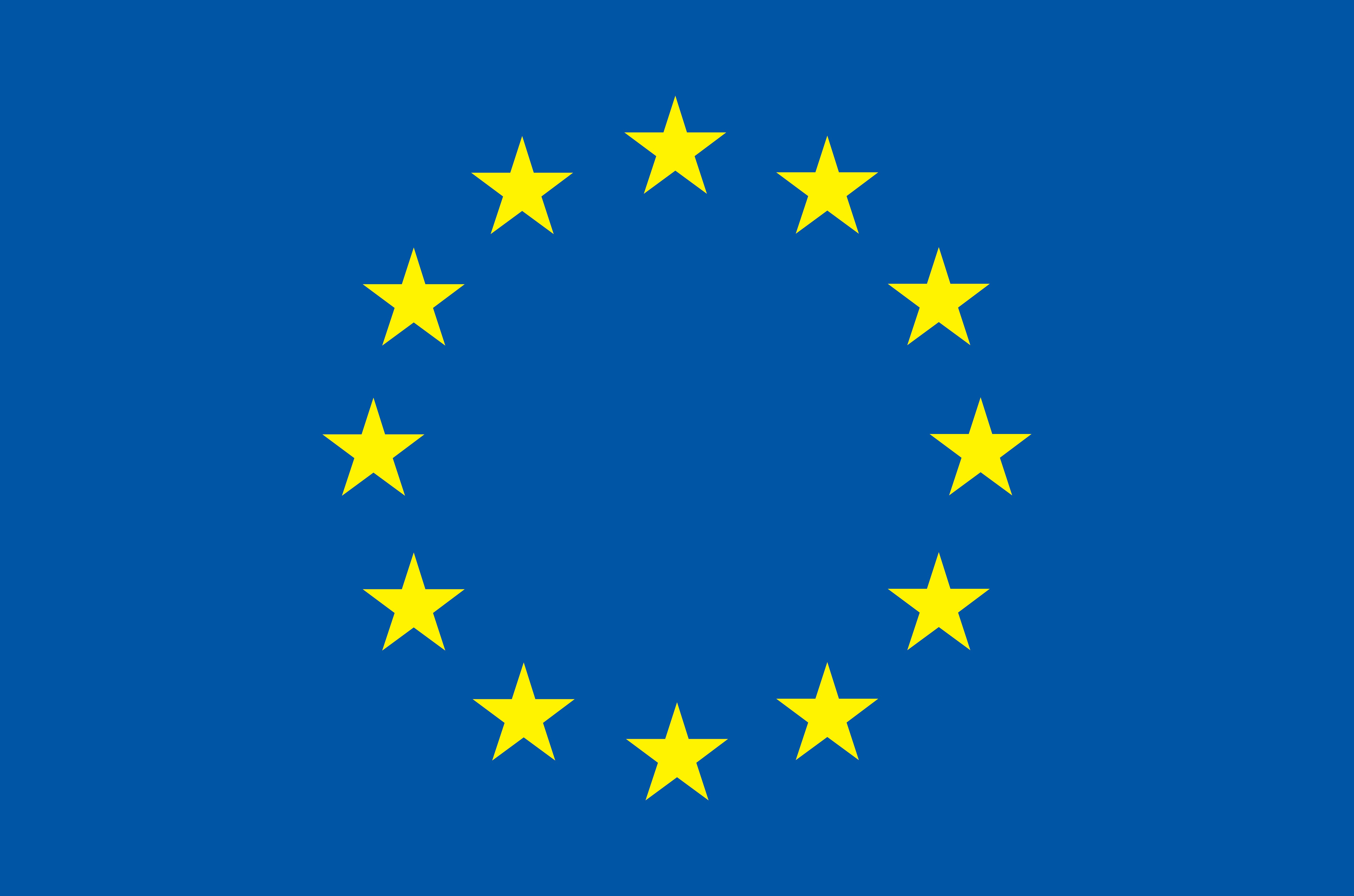} This project has received funding from the 
European Union’s Horizon 2020 research and innovation programme under grant 
agreement No 956123, and from UK Dstl under project SOLITUDE. GJ and XH are also partially supported by the UK EPSRC 
under projects 
[EP/R026173/1,~EP/T026995/1].



\clearpage
{\small
\bibliographystyle{ieee_fullname}
\bibliography{egbib}
}

\clearpage

\appendix
\section{Details of $\sigma_l$ and other $\Phi^{\adv}$}
\label{appendix:a}

\paragraph{Assumption of $\sigma_l$ in \cite{farnia2018generalizable}.} To prove \cref{eq:2} according to Theorem 1.5 in \cite{tropp2012user}, \cite{farnia2018generalizable} considers $f_{\widetilde{\W}}$ such that $\Big | ||\W_l||_2-||\widetilde{\W_l}||_2 \Big|\le \frac{1}{n}||\widetilde{\W_l}||_2$, then they assume $\sigma_l=\frac{||\widetilde{\W_l}||_2}{\beta_{\widetilde{\W_l}}}\sigma$, where $\beta_{\widetilde{\W_l}}:=\Big( \prod_{l=1}^n||\widetilde{\W_l}||_2 \Big)^\frac{1}{n}$.   

\paragraph{Assumption of $\sigma_l$ in \cite{neyshabur2017pac}.} To prove the PAC-Bayesian bound in \cite{neyshabur2017pac} according to Theorem 1.5 in \cite{tropp2012user}, \cite{neyshabur2017pac} assumes all variances are same across layers, that is, $\sigma_l=\sigma$.

\paragraph{Our assumption of $\sigma_l$.} We can prove Lem.~\ref{lemma:3.2} under both of above assumptions. To make the main paper more clear, we assume that $\sigma_l=\sigma$ in the main paper. And we provide the proofs of Lem.~\ref{lemma:3.2} for $\sigma_l=\sigma$ and $\sigma_l=\frac{||\widetilde{\W_l}||_2}{\beta_{\widetilde{\W_l}}}\sigma$ in \cref{appendix:b} (the assumption of $\sigma_l=\frac{||\widetilde{\W_l}||_2}{\beta_{\widetilde{\W_l}}}\sigma$ includes the assumption of $\sigma_l=\sigma$).

\paragraph{PGM attack for $\Phi^{\adv}$.} For a PGM attack with noise power $\epsilon$ given Euclidean norm $||\cdot||$, $r$ iterations for attack and step size $\mathcal{Z}$, let $\kappa\le ||\nabla_{\s''}\Lc(f_\W(\s''))||$ hold for every $\s''\in\{ \D \cup \D' \}$ with constant $\kappa>0$, then we get \cite{farnia2018generalizable} 
\begin{equation}
\begin{aligned}
    &\Phi^{\adv}=\Big\{\prod_{l=1}^n ||\W_l||_2 \Big( 1+\frac{\Z}{\kappa} \frac{1-(2\Z/\kappa)^r\overline{\lip} (\nabla \Lc\circ f_{\W})^r}{1-(2\Z/\kappa)\overline{\lip} (\nabla \Lc\circ f_{\W})}\\
    &\; (\prod_{l=1}^n||\W_l||_2) \sum_{l=1}^n\prod_{j=1}^l ||\W_j||_2 \Big) \Big\}^2 \sum_{l=1}^n \frac{||\W_l||^2_F}{||\W_l||^2_2},
\end{aligned}
\end{equation}
where 
\begin{equation}\nonumber
\overline{\lip} (\nabla \Lc\circ f_{\W}):=(\prod_{l=1}^n||\W_l||_2) \sum_{l=1}^n\prod_{j=1}^l ||\W_j||_2    
\end{equation}
gives an upper bound on the Lipschitz constant of $\nabla_{\s}\Lc(f_\W(\s))$.

\section{Proof of Lem.~\ref{lemma:3.2}}
\label{appendix:b}
\paragraph{We provide our proofs based on the proofs in \cite{farnia2018generalizable}, to be clearer about the proofs, we suggest readers go through Appendix C.2 in \cite{farnia2018generalizable} firstly.} 
To prove \cref{eq:2}, \cite{farnia2018generalizable} considers $f_{\widetilde{\W}}$ such that $\Big | ||\W_l||_2-||\widetilde{\W_l}||_2 \Big|\le \frac{1}{n}||\widetilde{\W_l}||_2$, since $(1+\frac{1}{n})^n\le e$ and $\frac{1}{e}\le (1-\frac{1}{n})^{n-1}$, we get 
\begin{equation}
    \Big(\frac{1}{e}\Big)^{\frac{n}{n-1}}\prod_{l=1}^n ||\widetilde{\W_l}||_2 \le \prod_{l=1}^n ||\W_l||_2 \le e \prod_{l=1}^n ||\widetilde{\W_l}||_2, 
\end{equation}
and for each $j$, we get 
\begin{equation}
\label{eq:25}
\begin{aligned}
    \frac{1}{||\W_j||_2}\prod_{l=1}^n ||\W_l||_2 \le \frac{e}{||\widetilde{\W_j}||_2}\prod_{l=1}^n ||\widetilde{\W_l}||_2 
\end{aligned}
\end{equation}
and
\begin{equation}
\label{eq:26}
\begin{aligned}
    \frac{1}{||\widetilde{\W_j}||_2}\prod_{l=1}^n ||\widetilde{\W_l}||_2 &\le (1-\frac{1}{n})^{-(n-1)}\frac{1}{||\W_j||_2}\prod_{l=1}^n ||\W_l||_2 \\
    &\le \frac{e}{||\W_j||_2}\prod_{l=1}^n ||\W_l||_2. 
\end{aligned}
\end{equation}

\noindent Then let $\sigma_l=\sigma$ ($\frac{||\widetilde{\W_l}||_2}{\beta_{\widetilde{\W_l}}}=1$) or $\sigma_l=\frac{||\widetilde{\W_l}||_2}{\beta_{\widetilde{\W_l}}}\sigma$ and let FGM perturbs vector be
\begin{equation}
    \delta^{\fgm}_{\W}(\s):= \mathop{\arg\max}\limits_{||\boldsymbol{\delta}||\le  \epsilon} \boldsymbol{\delta}^\T \nabla_{\s}\Lc(f_\W(\s)).
\end{equation}
According to Appendix C.2 Eq. (22) in \cite{farnia2018generalizable}, we get the following inequation
\begin{equation}
\label{eq:28}
\begin{aligned}
&||f_{\W+\U}\big(\s+\delta^{\fgm}_{\W+\U}(\s)\big)-f_\W\big(\s+\delta^{\fgm}_{\W}(\s)\big)||\\
&\le e(B+\epsilon)\prod_{l=1}^n||\W_l||_2 \sum_{l=1}^n \frac{||\U_l||_2}{||\W_l||_2}\\
&\quad+2e^2 \frac{\epsilon}{\kappa} \prod_{l=1}^n||\W_l||_2^2 \sum_{l=1}^n\Big[ \frac{||\U_l||_2}{||\W_l||_2}\\
&\quad\quad+B(\prod_{j=1}^l ||\W_j||_2) \sum_{j=1}^l \frac{||\U_j||_2}{||\W_j||_2} \Big].
\end{aligned}
\end{equation}

\noindent According to Section 1.1 in \cite{bandeira2021spectral}, we have
\begin{equation}\nonumber
\begin{aligned}
   \E||\U_l||_2&\lesssim (1+\sqrt{\ln h})||\E(\U_l^\T\U_l)||_2^{\frac{1}{2}}+||\E(\U_l\U_l^\T)||_2^{\frac{1}{2}} \\
   &\le c\Big( (1+\sqrt{\ln h})||\E(\U_l^\T\U_l)||_2^{\frac{1}{2}}+||\E(\U_l\U_l^\T)||_2^{\frac{1}{2}} \Big),
\end{aligned}
\end{equation}
\begin{equation}\nonumber
    \Pro\Big(\Big| ||\U_l||_2-\E||\U_l||_2 \Big|\ge t\Big)\le 2e^{-t^2/2\sigma_{*}(\U_l)^2},
\end{equation}
\begin{equation}\nonumber
    \sigma_{*}(\U_l)\le ||\E(\U_l^\T\U_l)||_2^{\frac{1}{2}},
\end{equation}
where $c>0$ is a universal constant.
Taking a union bond over the layers, we get that, with probability $>\frac{1}{2}$, the spectral norm of $\U_l$ is bounded by $(\sqrt{2\ln(4n)}+c+c\sqrt{\ln h})||\E(\U_l^\T\U_l)||_2^{\frac{1}{2}}+c||\E(\U_l\U_l^\T)||_2^{\frac{1}{2}}$, let $c_1=\sqrt{2\ln(4n)}+c+c\sqrt{\ln h}$ and $c_2=c$, we have
\begin{equation}
\label{eq:29}
||\U_l||_2 \le \Big ( c_1||\R'_l||_2^{\frac{1}{2}}+c_2||\R''_l||_2^{\frac{1}{2}} \Big)\sigma_l.
\end{equation}
Thus, $\frac{\beta_{\widetilde{\W_l}}}{||\widetilde{\W_l}||_2}||\U_l||_2$ is bounded by $\Big ( c_1||\R'_l||_2^{\frac{1}{2}}+c_2||\R''_l||_2^{\frac{1}{2}} \Big)\sigma$. 
Then, according to Appendix C.2 Eq. (22) in \cite{farnia2018generalizable}, \cref{eq:25,eq:28}, we can get
\begin{small}
\begin{equation}\nonumber
\begin{aligned}
&||f_{\W+\U}\big(\s+\delta^{\fgm}_{\W+\U}(\s)\big)-f_\W\big(\s+\delta^{\fgm}_{\W}(\s)\big)||\\
&\le e^2(B+\epsilon)\prod_{l=1}^n||\widetilde{\W_l}||_2 \sum_{l=1}^n \frac{||\U_l||_2}{||\widetilde{\W_l}||_2}\\
&\quad+2e^5 \frac{\epsilon}{\kappa} \prod_{l=1}^n||\widetilde{\W_l}||_2^2 \sum_{l=1}^n\Big[ \frac{||\U_l||_2}{||\widetilde{\W_l}||_2}\\
&\quad\quad+B(\prod_{j=1}^l ||\widetilde{\W_j}||_2) \sum_{j=1}^l \frac{||\U_j||_2}{||\widetilde{\W_j}||_2} \Big]\\
&\le 2e^5 (B+\epsilon) \sigma \Big (\sum_{l=1}^n (c_1||\R'_l||_2^{\frac{1}{2}}+c_2||\R''_l||_2^{\frac{1}{2}}) \Big)\\
&\Big\{ \prod_{l=1}^n ||\widetilde{\W_l}||_2^{\frac{n-1}{n}}+\frac{\epsilon}{\kappa}\Big(\prod_{l=1}^n ||\widetilde{\W_l}||_2^{\frac{2n-1}{n}}\Big) \Big( \frac{1}{B}+\sum_{l=1}^n \prod_{j=1}^l ||\widetilde{\W_j}||_2 \Big) \Big\}\\
&\le \frac{\gamma}{4},
\end{aligned}
\end{equation}
\end{small} hence we choose
\begin{small}
\begin{equation}
\begin{aligned}
&\sigma=\frac{\gamma}{8e^5(B+\epsilon)(\sum_{l=1}^n (c_1||\R'_l||_2^{\frac{1}{2}}+c_2||\R''_l||_2^{\frac{1}{2}}))\prod_{l=1}^n ||\widetilde{\W_l}||_2^{\frac{n-1}{n}}}\\
&\cdot \frac{1}{\big(1+\frac{\epsilon}{\kappa}\prod_{l=1}^n ||\widetilde{\W_l}||_2(\frac{1}{B}+\sum_{l=1}^n \prod_{j=1}^l ||\widetilde{\W_j}||_2) \big)}
\end{aligned}
\end{equation}
\end{small}
Then we can get
\begin{equation}
\label{eq:31}
\begin{aligned}
&\KL(Q_{\vecc(\W)+\ul}||P)=\sum_{l=1}^n \Big( \frac{||\W_l||_F^2}{2\sigma_l^2}-\ln \det \R_l \Big)\\
&\le \mathcal{O}\Bigg( (B+\epsilon)^2 \Big(\sum_{l=1}^n (c_1||\R'_l||_2^{\frac{1}{2}}+c_2||\R''_l||_2^{\frac{1}{2}})\Big)^2 \prod_{l=1}^n ||\widetilde{\W_l}||_2^2\\
&\frac{\big(1+\frac{\epsilon}{\kappa}\prod_{l=1}^n ||\widetilde{\W_l}||_2\sum_{l=1}^n \prod_{j=1}^l ||\widetilde{\W_j}||_2 \big)^2}{\gamma^2}\sum_{l=1}^n \frac{||\W_l||^2_F}{||\widetilde{\W_l}||^2_2}\\
&-\sum_{l=1}^n \ln \det \R_l \Bigg) \\
&\le \mathcal{O}\Bigg( (B+\epsilon)^2 \Big(\sum_{l=1}^n (c_1||\R'_l||_2^{\frac{1}{2}}+c_2||\R''_l||_2^{\frac{1}{2}})\Big)^2 \prod_{l=1}^n ||\W_l||_2^2\\
&\frac{\big(1+\frac{\epsilon}{\kappa}\prod_{l=1}^n ||\W_l||_2\sum_{l=1}^n \prod_{j=1}^l ||\W_j||_2 \big)^2}{\gamma^2}\sum_{l=1}^n \frac{||\W_l||^2_F}{||\W_l||^2_2}\\
&-\sum_{l=1}^n \ln \det \R_l \Bigg).
\end{aligned}
\end{equation}
Thus, we have 
\begin{small}
\begin{equation}\nonumber
\begin{aligned}
&\Lc_{\D'}(f_\W) \le  \Lc_{\gamma,\St'}(f_\W)+\mathcal{O} \Bigg( \bigg(\frac{-\sum_l \ln \det \R_l+\ln\frac{m}{\delta}}{\gamma^2m}\\
&+ \frac{\Psi^{\adv}\Big(\sum_l\big(c_1||\R'_l||_2^{\frac{1}{2}}+c_2||\R''_l||_2^{\frac{1}{2}}\big)\Big)^{2}}{\gamma^2m} \bigg)^{\frac{1}{2}} \Bigg),
\end{aligned}
\end{equation}
\end{small}where $\Psi^{\adv}=(B+\epsilon)^2\Phi^{\adv}$. 
And 
\begin{equation}
\begin{aligned}
    \Phi^{\adv} =&\prod_{l=1}^n ||\W_l||_2^2\Big\{ 1+\frac{\epsilon}{\kappa}(\prod_{l=1}^n||\W_l||_2)\\
    &\;\cdot \sum_{l=1}^n\prod_{j=1}^l ||\W_j||_2 \Big\}^2 \sum_{l=1}^n \frac{||\W_l||^2_F}{||\W_l||^2_2}
\end{aligned}
\end{equation}
for FGM attack. 

\noindent Proofs for PGM attack are similar (combine \cref{eq:29,eq:31} and Appendix C.3 in \cite{farnia2018generalizable}).

\section{Sampling Method}
\label{appendix:c}
\noindent We use sharpness-like method \cite{keskar2016large} to get a set of weight samples $(\W+\eta)$ such that $|\Lc(f_{\W+\eta})-\Lc(f_{\W})|\le \epsilon'$ (\emph{e.g.,} $\epsilon'=0.05$ for CIFAR-10/SVHN and $\epsilon'=0.1$ for CIFAR-100), where $\vecc(\eta)$ is a $\vec{0}$ mean Gaussian noise. To get the samples from the posteriori distribution steadily and fastly, we train the convergent network with learning rate 0.0001, noise $\eta$ and 50 epochs, then collect corresponding 50 samples. As the samples are stabilized at (clean/adversarial) training loss and validation loss but with different weights, we can treat them as the samples from same (clean/adversarial) posteriori distribution and estimate the correlation matrix through these samples.

\begin{figure*}[t!]
\includegraphics[width=1
\textwidth]{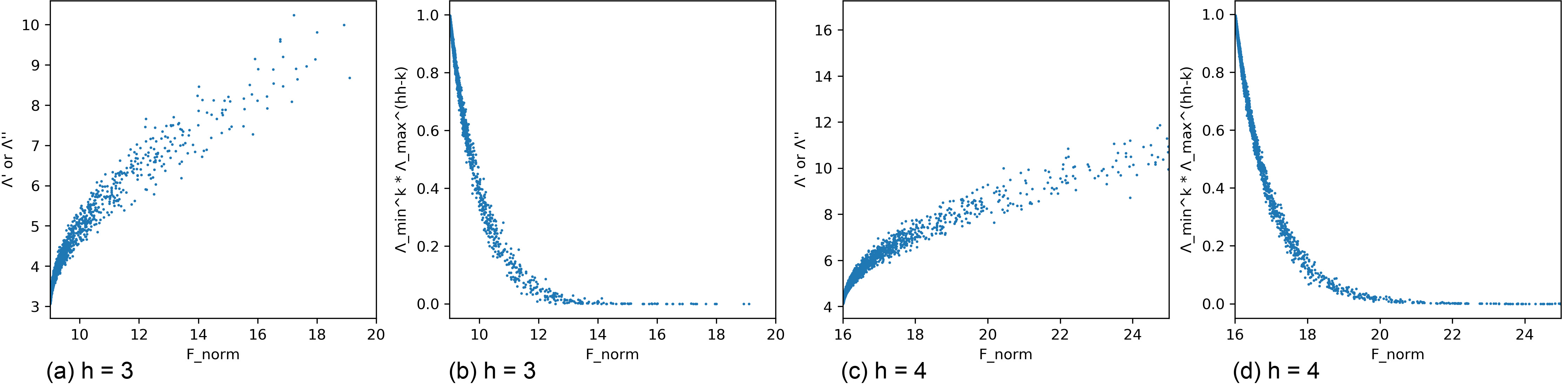}
\centering
\caption{\textbf{(a)} We sample 10000 9-dimensional correlation matrices and demonstrate $||\R_{l}||_F^2$ w.r.t $\Lambda'_{l,\max}$ or $\Lambda''_{l,\max}$. \textbf{(b)} We sample 10000 9-dimensional correlation matrices and demonstrate $||\R_{l}||_F^2$ w.r.t $\Lambda_{l,\min}^{k_l} \Lambda_{l,\max}^{\hh-k_l}$. \textbf{(c)} We sample 10000 16-dimensional correlation matrices and demonstrate $||\R_{l}||_F^2$ w.r.t $\Lambda'_{l,\max}$ or $\Lambda''_{l,\max}$. \textbf{(d)} We sample 10000 16-dimensional correlation matrices and demonstrate $||\R_{l}||_F^2$ w.r.t $\Lambda_{l,\min}^{k_l} \Lambda_{l,\max}^{\hh-k_l}$.}   
\label{fig:s}
\end{figure*}

\begin{figure*}[t!]
\includegraphics[width=1
\textwidth]{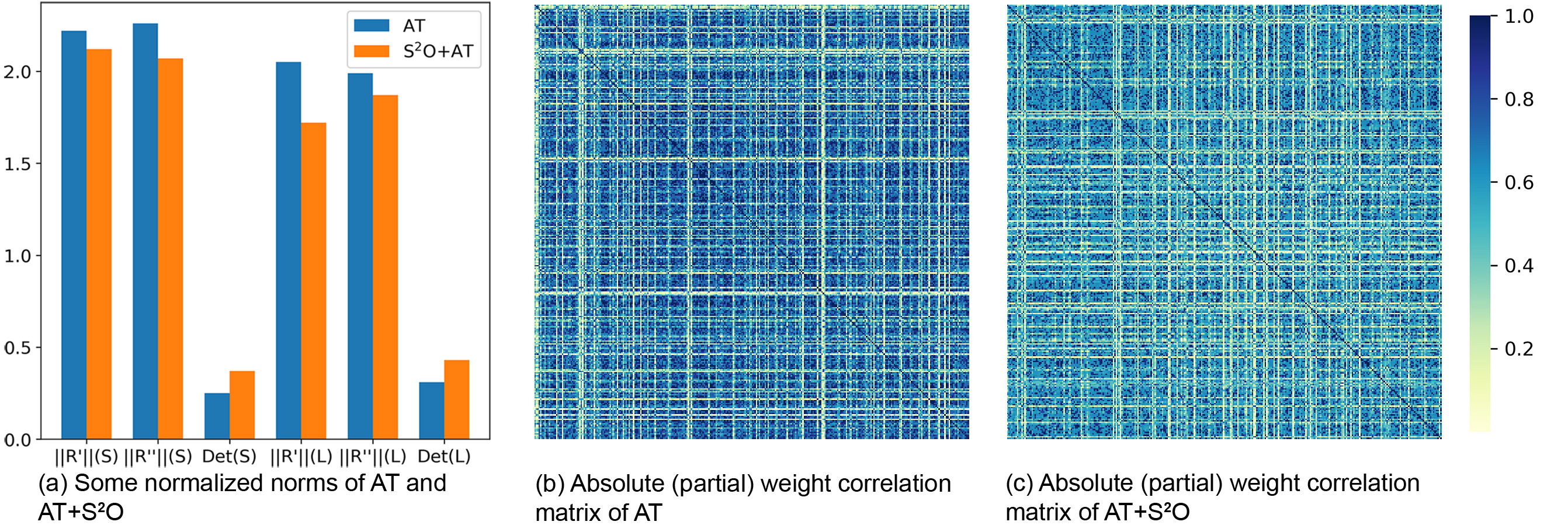}
\centering
\vspace{-5mm}
\caption{\textbf{(a)} shows the normalized spectral norm of $\R'_{\St}$, ${\R''}_{\St}$, and the determinant of $\R_{\St}$, with sampling estimation (S) and Laplace approximation (L) respectively. \textbf{(b)} and \textbf{(c)} demonstrate the absolute correlation matrix of partial weights (estimate under clean data), for AT and AT+S$^2$O respectively.}   
\vspace{-2mm}
\label{fig:s2}
\end{figure*}

\section{Proofs of Lems. \ref{lem:4.1}, \ref{lem:4.2}}
\label{appendix:d}
\noindent As we assume $r_\s r_{\s'}\ge 0$ (above Lem.~\ref{lem:4.1}), we give the proofs with two cases ($r_\s \ge 0$ and $r_\s \le 0$).

\paragraph{Proof for Lem. \ref{lem:4.1}.}
\quad

\noindent Let $r_{\s}\ge 0$ and $ r_{\s'}\ge 0$, we get
\begin{equation}
\begin{aligned}
\Lambda'_{l,\max}&=\max\big(||\R'_{l,\St}||_2^{\frac{1}{2}},||\R'_{l,\St'}||_2^{\frac{1}{2}}\big)\\
&=\sqrt{h\big(1+(h-1)\max(r_\s,r_{\s'})\big)}
\end{aligned}    
\end{equation}
and 
\begin{equation}
\begin{aligned}
\Lambda''_{l,\max}&=\max\big(||\R''_{l,\St}||_2^{\frac{1}{2}},||\R''_{l,\St'}||_2^{\frac{1}{2}}\big)\\
&=\sqrt{h\big(1+(h-1)\max(r_\s,r_{\s'})\big)}.
\end{aligned}    
\end{equation}
Thus, decreasing $||\R_{l,\St}||_F^2$ and $||\R_{l,\St'}||_F^2$ leads to a decline in  $\Lambda'_{l,\max}$ and $\Lambda''_{l,\max}$.

\quad

\noindent Let $r_{\s}\le 0$ and $ r_{\s'}\le 0$, we get
\begin{equation}
\begin{aligned}
\Lambda'_{l,\max}&=\max\big(||\R'_{l,\St}||_2^{\frac{1}{2}},||\R'_{l,\St'}||_2^{\frac{1}{2}}\big)\\
&=\sqrt{h\big(1-\min(r_\s,r_{\s'})\big)}
\end{aligned}    
\end{equation}
and 
\begin{equation}
\begin{aligned}
\Lambda''_{l,\max}&=\max\big(||\R''_{l,\St}||_2^{\frac{1}{2}},||\R''_{l,\St'}||_2^{\frac{1}{2}}\big)\\
&=\sqrt{h\big(1-\min(r_\s,r_{\s'})\big)}.
\end{aligned}    
\end{equation}
Thus, decreasing $||\R_{l,\St}||_F^2$ and $||\R_{l,\St'}||_F^2$ leads to a decline in  $\Lambda'_{l,\max}$ and $\Lambda''_{l,\max}$.

\paragraph{Proof for Lem. \ref{lem:4.2}.} 
\quad

\noindent Let $r_{\s}\ge r_{\s'}\ge 0$, we get
\begin{equation}
\begin{aligned}
    c(r)&=\Lambda_{l,\min}^{k_l} \Lambda_{l,\max}^{\hh-k_l}\\
    &=(1-r_\s)^{\hh-1}(1+(\hh-1)r_\s)\\
\end{aligned}
\end{equation}
and 
\begin{equation}
\begin{aligned}
    \frac{\partial c(r)}{\partial r_\s} = -\hh(\hh-1)r_\s (1-r_\s)^{\hh-2}\le 0,
\end{aligned}
\end{equation}
it is easy to get $c(r)$ is negative correlated with $r_\s$. Similarly, if $r_{\s'}\ge r_{\s}\ge 0$, we can get $c(r)$ is negative correlated with $r_{\s'}$. Thus, decreasing $||\R_{l,\St}||_F^2$ and $||\R_{l,\St'}||_F^2$ leads to an increase in $\Lambda_{l,\min}^{k_l} \Lambda_{l,\max}^{\hh-k_l}$.  

\quad

\noindent Let $r_{\s}\le r_{\s'}\le 0$, we get
\begin{equation}
\begin{aligned}
    c(r)&=\Lambda_{l,\min}^{k_l} \Lambda_{l,\max}^{\hh-k_l}\\
    &=(1+(\hh-1)r_\s)(1-r_\s)^{\hh-1}\\
\end{aligned}
\end{equation}
and
\begin{equation}
\begin{aligned}
    \frac{\partial c(r)}{\partial r_\s} = -\hh(\hh-1)r_\s (1-r_\s)^{\hh-2}\ge 0,
\end{aligned}
\end{equation}
it is also easy to get $c(r)$ is positive correlated with $r_\s$. Similarly, if $r_{\s'}\le r_{\s}\le 0$, we can get $c(r)$ is positive correlated with $r_{\s'}$. Thus, decreasing $||\R_{l,\St}||_F^2$ and $||\R_{l,\St'}||_F^2$ leads to an increase in $\Lambda_{l,\min}^{k_l} \Lambda_{l,\max}^{\hh-k_l}$.

\section{Simulations of Lems.~\ref{lem:4.1}, \ref{lem:4.2} and Second-Order Statistics of Weights under Clean Data}
\label{appendix:e}

\noindent As \cref{fig:s} shows, for 10000 random general 9-dimensional correlation matrices and 16-dimensional correlation matrices respectively, Lems.~\ref{lem:4.1} and \ref{lem:4.2} also hold approximately.

\quad

\noindent The results in \cref{fig:s2} also suggest that S$^2$O can decrease the spectral norm of $\R'_{\St}$, ${\R''}_{\St}$ and increases the determinant of $\R_{\St}$.

\section{Approximate Optimization}
\label{appendix:f}
We use a fast approximate method to update $\g(\A)$, i.e., add a penalty term to the high correlated $\ac_{l,i}$ and $\ac_{l,j}$ to reduce their correlation. Details are given in the code.

\end{document}